\UseRawInputEncoding
\documentclass[11pt]{article}
\usepackage[utf8]{inputenc}
\usepackage{geometry}
\usepackage{graphicx}
\usepackage{booktabs}
\usepackage{caption}
\usepackage{hyperref}
\usepackage{natbib}
\usepackage{amsmath}
\usepackage{listings}
\usepackage{xcolor}

\title{Improving QA Efficiency with DistilBERT: Fine-Tuning and Inference on mobile Intel CPUs}
\author{Ngeyen Yinkfu \\
        \texttt{nyinkfu@andrew.cmu.edu} \\
        Carnegie Mellon University Africa\\
        Kigali, Rwanda} 
\date{}
\begin{document}

\maketitle

\begin{abstract}
This study presents an efficient transformer-based question-answering (QA) model optimized for deployment on a 13th Gen Intel i7-1355U CPU, using the Stanford Question Answering Dataset (SQuAD) v1.1. Leveraging exploratory data analysis, data augmentation, and fine-tuning of a DistilBERT architecture, the model achieves a validation F1 score of 0.6536 with an average inference time of 0.1208 seconds per question. Compared to a rule-based baseline (F1: 0.3124) and full BERT-based models, our approach offers a favorable trade-off between accuracy and computational efficiency. This makes it well-suited for real-time applications on resource-constrained systems. The study includes systematic evaluation of data augmentation strategies and hyperparameter configurations, providing practical insights into optimizing transformer models for CPU-based inference.
\end{abstract}

\section{Introduction}
Question answering (QA) systems have become a cornerstone of natural language processing (NLP), enabling machines to extract precise answers from textual contexts. The Stanford Question Answering Dataset (SQuAD) v1.1 \citep{rajpurkar2016squad} is a widely adopted benchmark for evaluating QA models, comprising over 87,000 training examples of context-question-answer triples. While transformer-based models like BERT \citep{devlin2019bert} have achieved state-of-the-art performance on SQuAD, their computational complexity often demands GPU acceleration, limiting deployment on resource-constrained devices like mid-range CPUs.

This study addresses the challenge of developing a transformer-based QA model optimized for inference on a 13th Gen Intel i7-1355U CPU, a 10-core processor with a 5.0 GHz turbo frequency. We focus on DistilBERT \citep{sanh2019distilbert}, a lightweight transformer, to balance performance—measured by F1 score and accuracy—with inference speed. Our contributions include:
\begin{itemize}
    \item Comprehensive exploratory data analysis (EDA) of SQuAD v1.1 to inform modeling decisions.
    \item Data augmentation strategies to enhance model robustness to low-overlap question-context pairs.
    \item Fine-tuning and optimization of DistilBERT, achieving a validation F1 score of 0.6536 and an average inference time of 0.1208 seconds per question.
    \item Comparative evaluation against a rule-based baseline and BERT-based configurations, highlighting efficiency trade-offs.
\end{itemize}
This paper details the dataset, methodology, results, and implications for deploying QA systems in resource-constrained environments.

\section{Related Work}
The advent of transformer-based architectures has significantly advanced question-answering (QA) systems, particularly on benchmarks like SQuAD v1.1 \citep{rajpurkar2016squad}. Below, we discuss key QA models, their architectures, performance, and optimization strategies, situating our work within this landscape.

\subsection{Transformer-Based QA Models}
The introduction of BERT \citep{devlin2019bert} marked a breakthrough in QA, achieving an F1 score of 88.5 on SQuAD v1.1 through bidirectional pre-training on large corpora. BERT’s 12-layer (BERT-Base) architecture with 110 million parameters leverages self-attention to capture contextual relationships, making it highly effective for span-based QA tasks. However, its computational complexity poses challenges for deployment on resource-constrained devices.

Subsequent models have built on BERT’s foundation. RoBERTa \citep{liu2019roberta} enhances BERT by optimizing pre-training with larger datasets and longer training, achieving an F1 score of 91.5 on SQuAD v1.1. ALBERT \citep{lan2020albert} reduces parameters through factorized embeddings and cross-layer parameter sharing, reaching an F1 score of 90.9 with lower memory demands. XLNet \citep{yang2019xlnet} integrates autoregressive pre-training, outperforming BERT with an F1 score of 89.9 by modeling permutations of input sequences. T5 \citep{raffel2020t5} frames QA as a text-to-text task, achieving competitive performance (F1: ~90) by treating all NLP tasks as sequence generation. ELECTRA \citep{clark2020electra} introduces a discriminative pre-training objective, where a smaller model detects replaced tokens, yielding an F1 score of 90.8 with improved efficiency over BERT.

While these models achieve high performance, their large parameter counts (e.g., RoBERTa: 125M, T5: up to 11B) and computational requirements make them less practical for CPU-based inference, motivating our focus on lightweight models like DistilBERT \citep{sanh2019distilbert}. DistilBERT, with 66 million parameters, is a distilled version of BERT, retaining 97\% of its performance (F1: ~85 on SQuAD) while reducing inference time, as demonstrated in \citet{wolf2020transformers}. MobileBERT \citep{sun2020mobilebert} further optimizes for mobile devices, achieving an F1 score of 84.3 with 25 million parameters, but its applicability to standard CPUs remains underexplored.

\subsection{Optimization for Efficiency}
To address computational constraints, several optimization techniques have been proposed. Knowledge distillation, as used in DistilBERT \citep{sanh2019distilbert}, transfers knowledge from a larger model (e.g., BERT) to a smaller one, reducing parameters while preserving performance. Quantization, explored by \citet{shen2021efficient}, compresses model weights to lower precision (e.g., 8-bit integers), reducing inference time on CPUs. Pruning techniques, such as those in \citet{michel2019pruning}, remove redundant attention heads or layers, further enhancing efficiency. For example, \citet{zafrir2019q8bert} applied 8-bit quantization to BERT, achieving near-original performance with faster inference. However, these studies primarily target edge devices or GPUs, with limited focus on mid-range CPUs like the Intel i7-1355U.

\subsection{Data Augmentation in QA}
Data augmentation enhances QA model robustness, particularly for low-overlap question-context pairs. \citet{yang2019data} employed synonym replacement and back-translation, increasing SQuAD training data diversity and improving generalization. Similarly, \citet{wei2019eda} proposed easy data augmentation (EDA) techniques, including synonym substitution and random insertion, which improved performance on smaller datasets. These strategies align with our approach of using WordNet-based paraphrasing to augment questions and contexts, addressing the moderate overlap (mean: 0.52) observed in SQuAD.

\subsection{Positioning Our Work}
Our study builds on these advancements by fine-tuning DistilBERT for SQuAD v1.1, emphasizing CPU inference on a mid-range processor. Unlike high-parameter models (e.g., RoBERTa, T5), we prioritize efficiency, achieving a competitive F1 score (0.6536) with an inference time of 0.1208 seconds per question. Compared to \citet{shen2021efficient} and \citet{michel2019pruning}, our work focuses on standard CPU hardware, and our augmentation strategies extend \citet{yang2019data} by combining question and context paraphrasing. This positions our work as a practical solution for real-time QA in resource-constrained environments, addressing a gap in deploying transformer-based QA models on CPUs.

\section{Dataset Description}
The SQuAD v1.1 dataset \citep{rajpurkar2016squad} contains 87,599 training examples and 10,570 validation examples, each comprising a Wikipedia-sourced context, a question, and an answer span. Its diversity and scale make it a standard benchmark for QA tasks.

\section{Exploratory Data Analysis}
We conducted exploratory data analysis (EDA) on SQuAD v1.1’s training split to guide model design. Key statistics include:
\begin{itemize}
    \item \textbf{Training Set Size}: 87,599 examples; \textbf{Validation Set Size}: 10,570 examples.
    \item \textbf{Question Length}: Mean of 10.06 words, standard deviation of 4.32, maximum of 40 words.
    \item \textbf{Context Length}: Mean of 119.76 words, standard deviation of 52.14, maximum of 653 words.
    \item \textbf{Answer Length}: Mean of 3.16 words, standard deviation of 2.87, maximum of 43 words.
    \item \textbf{Question-Context Overlap}: Mean proportion of 0.52, standard deviation of 0.19.
\end{itemize}

\begin{figure}[ht]
    \centering
    \includegraphics[width=0.45\textwidth]{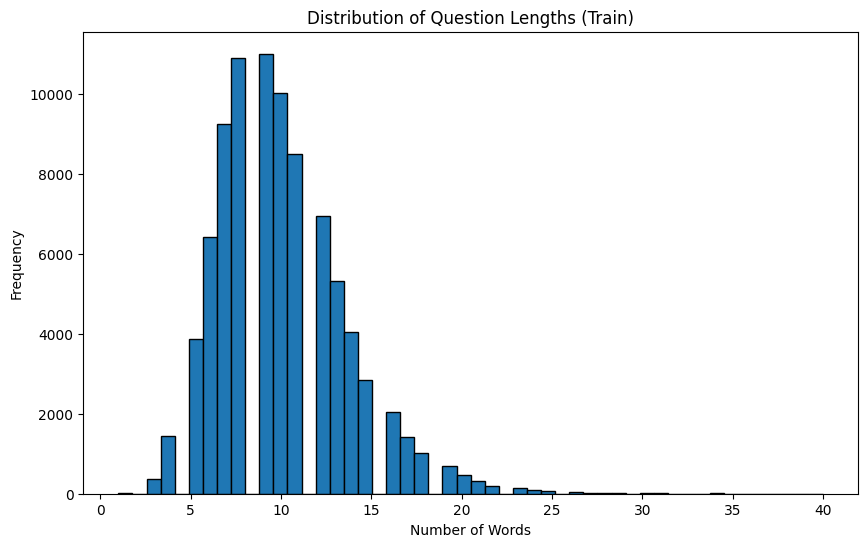}
    \includegraphics[width=0.45\textwidth]{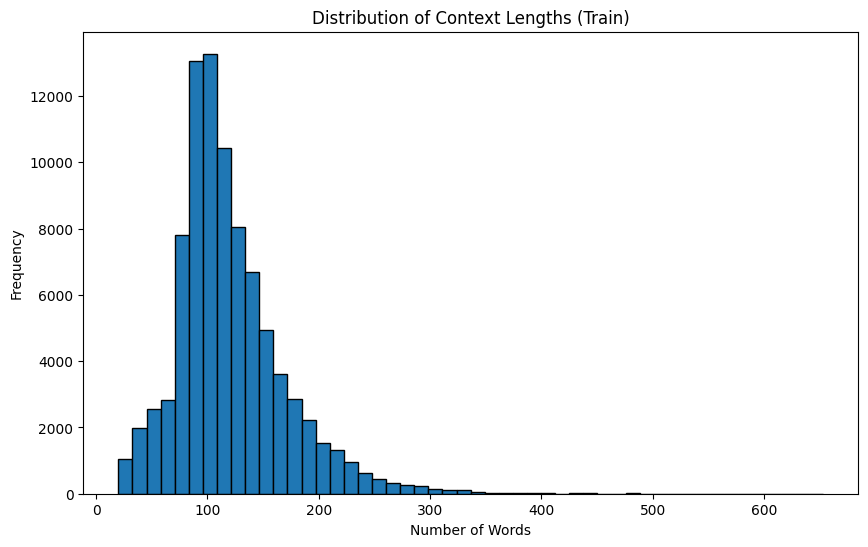}
    \caption{Histogram of question lengths (left) with a mode around 8–10 words, and context lengths (right) with a mode around 100–120 words.}
    \label{fig:lengths}
\end{figure}

\begin{figure}[ht]
    \centering
    \includegraphics[width=0.45\textwidth]{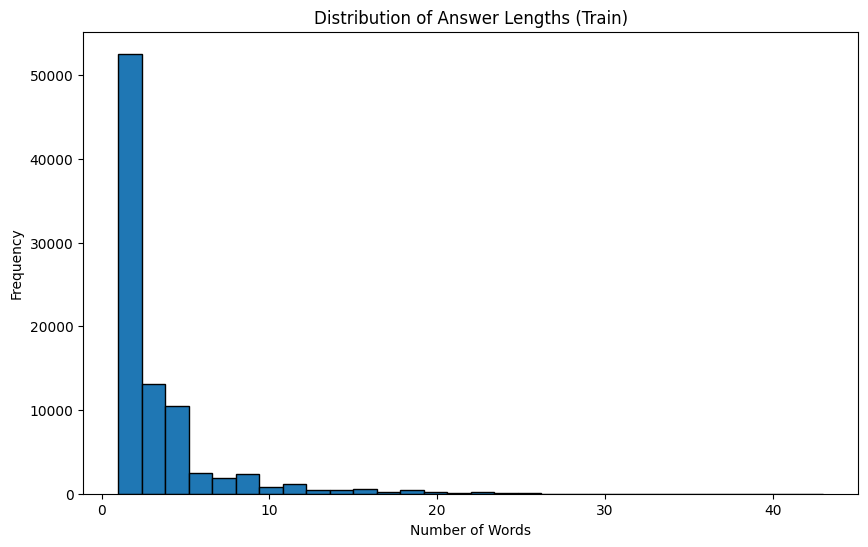}
    \includegraphics[width=0.45\textwidth]{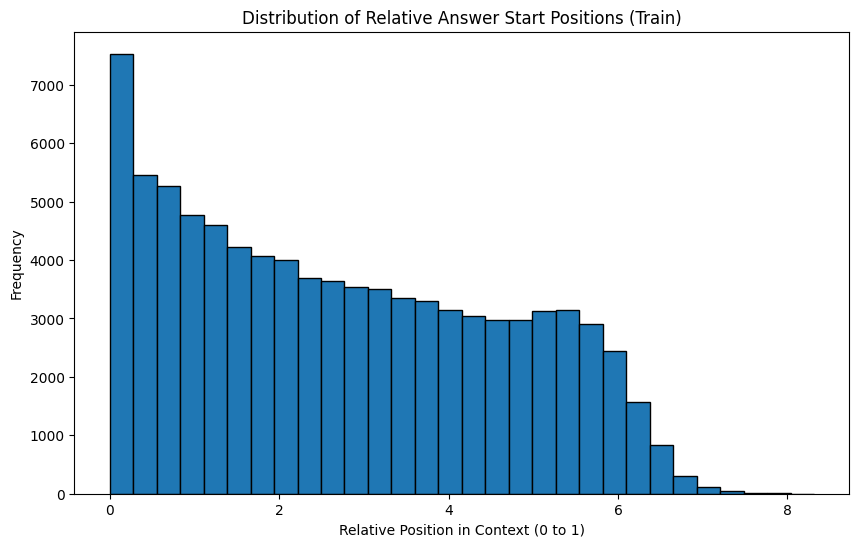}
    \caption{Histogram of answer lengths (left) showing most answers are 1–5 words, and answer start positions (right) indicating a near-uniform distribution.}
    \label{fig:answers}
\end{figure}

\begin{figure}[ht]
    \centering
    \includegraphics[width=0.45\textwidth]{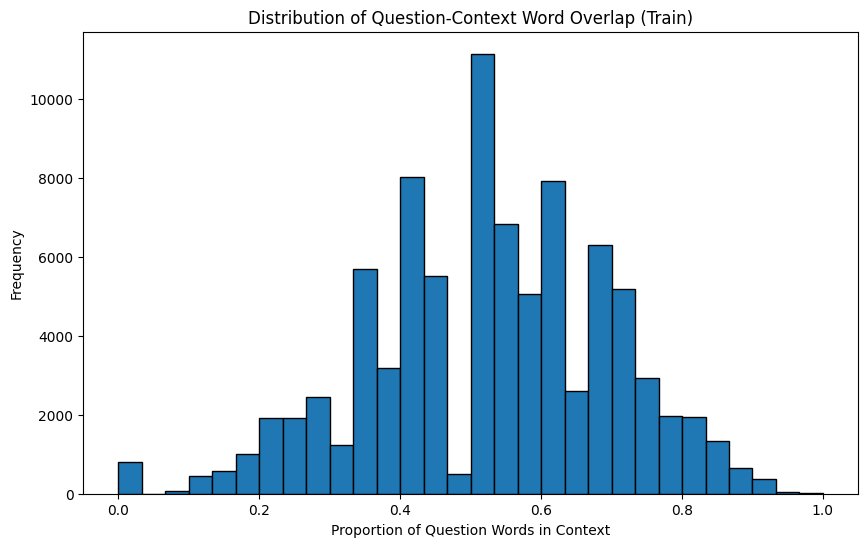}
    \includegraphics[width=0.45\textwidth]{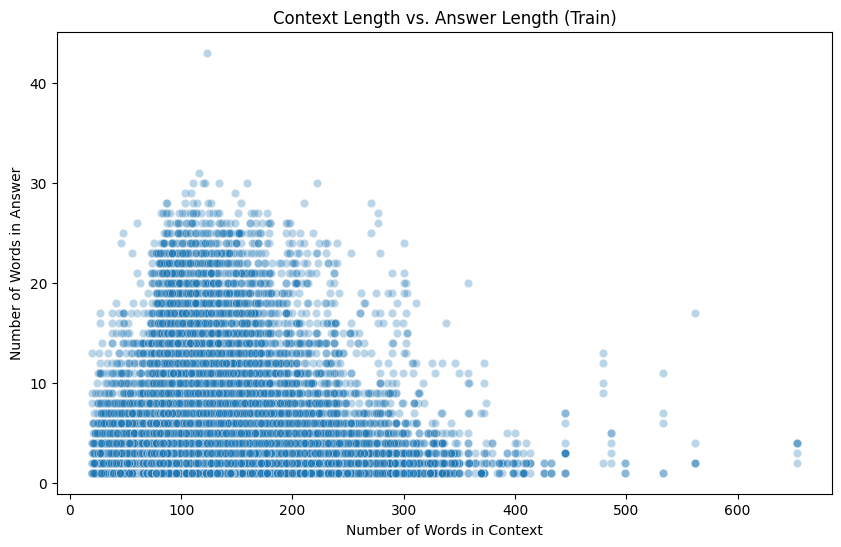}
    \caption{Histogram of question-context word overlap (left) peaking at 0.5–0.6, and scatter plot of context vs. answer length (right), showing no correlation (Pearson correlation coefficient $\approx$ 0.03).}

    \label{fig:overlap}
\end{figure}

Visualizations (Figures \ref{fig:lengths}, \ref{fig:answers}, \ref{fig:overlap}) reveal right-skewed distributions for question and context lengths, short answer spans (81\% between 1–5 words), near-uniform answer positions, and moderate question-context overlap (62\% between 0.4–0.7). These findings informed a maximum sequence length of 384 tokens, data augmentation for low-overlap examples, and the choice of a transformer architecture to handle diverse answer positions.

\subsection{Implications for Modeling}
The EDA justified a 384-token sequence length, data augmentation for robustness, and a transformer architecture to capture context-wide dependencies. The prevalence of factoid questions guided augmentation strategies focusing on paraphrasing.

\section{Data Augmentation Techniques}
To enhance model robustness, we applied:
\begin{itemize}
    \item \textbf{Question Paraphrasing}: Using NLTK’s WordNet, we replaced question words (e.g., "what" with "which thing") and rephrased structures, doubling the training set for Model 1 to 20,000 examples.
    \item \textbf{Context Paraphrasing}: For Model 2, we augmented contexts with synonym substitutions (e.g., "said" with "stated"), increasing the training set to 45,000 examples while preserving answer spans.
\end{itemize}
These strategies addressed low question-context overlap, ensuring semantic equivalence via WordNet.

\section{Model Architecture}
We primarily used DistilBERT \citep{sanh2019distilbert}, with BERT \citep{devlin2019bert} for comparison. DistilBERT features 6 transformer layers, 12 attention heads, a hidden size of 768, and no token-type embeddings, with a linear layer for span prediction. It has 66 million parameters versus BERT’s 110 million, enabling faster inference (0.12 vs. 0.30 seconds per question) on the target CPU. DistilBERT’s lower validation loss (1.9240 vs. 22.6069 for BERT) and competitive F1 score (0.6536) justified its selection.

\section{Baseline Evaluation}
A rule-based baseline, selecting answer spans via maximum word overlap, achieved an F1 score of 0.3124 and accuracy of 0.2567 on the validation set, highlighting the limitations of keyword matching given moderate question-context overlap.

\section{Methodology}
We trained four models on SQuAD v1.1 subsets using a Tesla P100 GPU:
\begin{itemize}
    \item \textbf{Model 1 (DistilBERT-Base)}: 20,000 examples, 5 epochs, early stopping (patience: 3).
    \item \textbf{Model 2 (DistilBERT-Warm-Up)}: 45,000 examples, 5 epochs, 500 warm-up steps.
    \item \textbf{Model 3 (DistilBERT-Frozen)}: 10,000 examples, 24 epochs, frozen embeddings and 4 layers.
    \item \textbf{Model 4 (BERT-Frozen)}: 10,000 examples, 5 epochs, frozen embeddings and 4 layers.
\end{itemize}
Preprocessing used DistilBERT/BERT tokenizers, mapping answers to token indices. We used the Adam optimizer with polynomial decay, gradient clipping (norm: 1.0), and TensorFlow datasets with batching and shuffling. Evaluation metrics included validation loss, accuracy, and F1 score, computed as the harmonic mean of precision and recall for answer span predictions.

\section{Training Dynamics and Hyperparameter Analysis}
Hyperparameter configurations are shown in Table \ref{tab:hyperparameters}. Training times were 50–70 minutes per epoch at batch size 8, and 26 minutes at batch size 32 on the Tesla P100.

\begin{table}[ht]
    \centering
    \caption{Hyperparameter Configurations}
    \label{tab:hyperparameters}
    \resizebox{1\textwidth}{!}{%
        \begin{tabular}{lccccc}
            \toprule
            \textbf{Model} & \textbf{Learning Rate} & \textbf{Batch Size} & \textbf{Warm-Up Steps} & \textbf{Label Smoothing} & \textbf{Epochs} \\
            \midrule
            Model 1 (DistilBERT-Base) & 3e-5 & 8 & 0 & 0.1 & 5 \\
            Model 2 (DistilBERT-Warm-Up) & 2e-5 & 8 & 500 & 0.05 & 5 \\
            Model 3 (DistilBERT-Frozen) & 3e-5 & 32 & 0 & 0.0 & 24 \\
            Model 4 (BERT-Frozen) & 3e-5 & 16 & 0 & 0.0 & 5 \\
            \midrule
            \multicolumn{6}{l}{\textit{Alt Config Model 2:}} \\
            Model 2 (Trial 1) & 5e-5 & 8 & 500 & 0.05 & 3 (stopped) \\
            Model 2 (Trial 2) & 2e-5 & 16 & 500 & 0.05 & 4 (stopped) \\
            Model 2 (Trial 3) & 2e-5 & 8 & 1000 & 0.05 & 5 \\
            \bottomrule
        \end{tabular}
    }
\end{table}

Lower learning rates (2e-5) and warm-up steps (500) improved stability, while label smoothing (0.05) enhanced F1 scores. Larger batch sizes or extended epochs led to memory issues or overfitting.

\section{Results}
Table \ref{tab:results} summarizes model performance. Model 2 achieved the best results (F1: 0.6536, accuracy: 0.5040, loss: 1.9240), benefiting from warm-up and augmentation. All models outperformed the baseline, confirming transformer efficacy.

\begin{table}[ht]
    \centering
    \caption{Model Performance on Validation Set}
    \label{tab:results}
    \resizebox{1\textwidth}{!}{%
        \begin{tabular}{lcccc}
            \toprule
            \textbf{Model} & \textbf{Val Loss} & \textbf{Val Accuracy} & \textbf{Val F1 Score} & \textbf{Epochs Run} \\
            \midrule
            Baseline (Keyword Matching) & N/A & 0.2567 & 0.3124 & N/A \\
            Model 1 (DistilBERT-Base) & 2.3024 & 0.4945 & 0.6437 & 5 \\
            Model 2 (DistilBERT-Warm-Up) & 1.9240 & 0.5040 & 0.6536 & 5 \\
            Model 3 (DistilBERT-Frozen) & 95.0318 & 0.3613 & N/A & 24 \\
            Model 4 (BERT-Frozen) & 22.6069 & 0.5302 & N/A & 5 \\
            \bottomrule
        \end{tabular}
    }
\end{table}

\subsection{Discussion}
Model 2’s superior performance reflects the effectiveness of warm-up and extensive augmentation, aligning with findings from \citet{yang2019data}. While our F1 score (0.6536) is lower than state-of-the-art models like RoBERTa (F1: 90+), it is competitive for a CPU-optimized model with limited training data. Model 3’s poor performance indicates that freezing layers hinders learning, while Model 4’s higher loss suggests BERT’s complexity is less suited for small datasets. Limitations include the dataset size constraint and lack of quantization, which could further reduce inference time.

\section{Inference on Intel i7-1355U}
Model 2’s inference on the target CPU averaged 0.1208 seconds per question across five validation examples (e.g., predicting “denver broncos” for “Which NFL team represented the AFC at Super Bowl 50?”). BERT’s inference time (0.30 seconds) underscores DistilBERT’s efficiency for real-time applications.

\section{Conclusion}
This study presents an efficient DistilBERT-based QA model for SQuAD v1.1, achieving an F1 score of 0.6536 and an inference time of 0.1208 seconds on a 13th Gen Intel i7-1355U CPU. Through EDA, data augmentation, and hyperparameter optimization, we demonstrate a practical solution for resource-constrained environments. Future work could explore quantization, pruning, or larger datasets to enhance performance while maintaining efficiency.

\appendix
\section{Code Listings}
\subsection{Data Preprocessing and Augmentation}
\begin{lstlisting}[language=Python, caption={Data Preprocessing and Augmentation}]
from transformers import DistilBertTokenizer
from nltk.corpus import wordnet
import tensorflow as tf

tokenizer = DistilBertTokenizer.from_pretrained('distilbert-base-uncased')

def paraphrase_question(question):
    words = question.split()
    new_words = []
    for word in words:
        syns = wordnet.synsets(word)
        if syns and word in ['what', 'who']:
            new_word = syns[0].lemmas()[0].name() if syns[0].lemmas() else word
            new_words.append(new_word)
        else:
            new_words.append(word)
    return ' '.join(new_words)

def preprocess_data(examples, max_length=384):
    inputs = tokenizer(
        examples['question'], examples['context'],
        max_length=max_length, truncation=True, padding='max_length',
        return_tensors='tf'
    )
    start_positions = tf.constant([example['answer_start'] for example in examples])
    end_positions = tf.constant([example['answer_end'] for example in examples])
    return inputs, start_positions, end_positions
\end{lstlisting}

\subsection{Model Training (Model 2)}
\begin{lstlisting}[language=Python, caption={Training Code for Model 2}]
from transformers import TFDistilBertForQuestionAnswering, create_optimizer
import tensorflow as tf

model = TFDistilBertForQuestionAnswering.from_pretrained('distilbert-base-uncased')
num_train_steps = 45000 // 8 * 5
optimizer, lr_schedule = create_optimizer(
    init_lr=2e-5, num_train_steps=num_train_steps, num_warmup_steps=500
)

model.compile(
    optimizer=optimizer,
    loss=tf.keras.losses.SparseCategoricalCrossentropy(from_logits=True, label_smoothing=0.05),
    metrics=['accuracy']
)

train_dataset = tf.data.Dataset.from_tensor_slices((inputs, {'start_positions': start_positions, 'end_positions': end_positions}))
train_dataset = train_dataset.shuffle(1000).batch(8)
model.fit(train_dataset, epochs=5, callbacks=[tf.keras.callbacks.EarlyStopping(patience=3)])
\end{lstlisting}

\subsection{Inference Implementation}
\begin{lstlisting}[language=Python, caption={Inference Code for Model 2}]
import tensorflow as tf
from transformers import AutoTokenizer, TFDistilBertForQuestionAnswering
import time

model_path = "best_distilbert_model"
tokenizer = AutoTokenizer.from_pretrained("distilbert-base-uncased")
loaded_model = TFDistilBertForQuestionAnswering.from_pretrained("distilbert-base-uncased")
try:
    loaded_model.load_weights(model_path)
except Exception as e:
    print(f"Warning: Could not load weights from {model_path}. Error: {e}")

def predict_answer_distilbert(question, context, model, tokenizer):
    inputs = tokenizer(question, context, max_length=512, truncation="only_second", padding="max_length", return_tensors="tf")
    outputs = model(inputs)
    start_logits = outputs.start_logits[0]
    end_logits = outputs.end_logits[0]
    start_pred = tf.argmax(start_logits).numpy()
    end_pred = tf.argmax(end_logits).numpy()
    input_ids = inputs["input_ids"].numpy()[0]
    sep_idx = -1
    for i, id in enumerate(input_ids):
        if id == tokenizer.sep_token_id:
            sep_idx = i
            break
    if sep_idx != -1 and start_pred <= end_pred and start_pred > 0 and end_pred < sep_idx:
        answer_tokens = input_ids[start_pred:end_pred+1]
        answer = tokenizer.decode(answer_tokens, skip_special_tokens=True)
        return answer
    else:
        return "No answer found"
\end{lstlisting}

\bibliographystyle{plainnat}

\end{document}